\pdfoutput=1

\documentclass[11pt]{article}

\usepackage{acl}

\usepackage{times}
\usepackage{latexsym}

\usepackage[T1]{fontenc}

\usepackage[utf8]{inputenc}

\usepackage{microtype}
\usepackage{lipsum} 
\usepackage{booktabs}
\usepackage{graphicx}
\usepackage{amsmath}
\usepackage[framemethod=tikz]{mdframed}
\usepackage{comment}
\usepackage{multirow}
\usepackage{hyperref}
\usepackage{adjustbox}
\usepackage{footmisc}
\usepackage[misc]{ifsym}
\usepackage{caption}
\usepackage{subcaption}
\usepackage{extarrows}
\usepackage{amsmath}
\usepackage{mathrsfs}
\usepackage{microtype}
\usepackage{color}
\usepackage{colortbl} 
\usepackage{mathrsfs}
\usepackage{float}
\usepackage{multicol}
\usepackage{array}
\usepackage{nicematrix}
\usepackage{longtable}
\usepackage{rotating}
\usepackage{pdflscape}
\usepackage{amsfonts}
\usepackage{tabularx}
\usepackage{lipsum} 
\usepackage{csquotes}
\usepackage{xcolor}

\usepackage{tikz}
\usetikzlibrary{bayesnet}

\usepackage{bm}

\title{
Deciphering the Interplay of Parametric and Non-parametric Memory in Retrieval-augmented Language Models
}

\author{Mehrdad Farahani\hspace{5mm}Richard Johansson \\
  Chalmers University of Technology and University of Gothenburg \\
  \texttt{\{mehrdad.farahani, richajo\}@chalmers.se}}

\begin{document}
\maketitle
\begin{abstract}
Generative language models often struggle with specialized or less-discussed knowledge. A potential solution is found in Retrieval-Augmented Generation (RAG) models which act like retrieving information before generating responses. In this study, we explore how the \textsc{Atlas} approach, a RAG model, decides between what it already knows (parametric) and what it retrieves (non-parametric).
We use causal mediation analysis and controlled experiments to examine how internal representations influence information processing. 
Our findings disentangle the effects of parametric knowledge and the retrieved context. They indicate that in cases where the model can choose between both types of information (parametric and non-parametric), it relies more on the context than the parametric knowledge. Furthermore, the analysis investigates the computations involved in \emph{how} the model uses the information from the context. We find that multiple mechanisms are active within the model and can be detected with mediation analysis: first, the decision of \emph{whether the context is relevant}, and second, how the encoder computes output representations to support copying when relevant.\footnote{The code used in this project is available at our GitHub repository: \href{https://github.com/m3hrdadfi/rag-memory-interplay}{github.com/m3hrdadfi/rag-memory-interplay}.}

\end{abstract}

\section{Introduction}

Natural Language Processing (NLP) has made significant progress in recent years, mostly because of the development of Large Language Models (LLMs). These models can perform a variety of tasks with minimal supervision. While pure generative LLMs are often capable of answering basic factual questions \cite{petroni2019language}, simply by reciting information memorized from their training sets and stored in the model parameters, they are much less reliable in scenarios that require more specialized knowledge that is discussed less frequently on the web \cite{kandpal2023large}.

RAG models combine retrieval-based and generative approaches and have been proposed as a way to address some of the drawbacks of basic generative language models in information-seeking scenarios. They improve the factual accuracy of answers to low-frequency queries \cite{kandpal2023large} as well as prediction consistency \cite{hagstrom2023effect}. These models employ a retriever to gather relevant external information and a generator to produce responses. This duality helps models generate text using both internal knowledge stored in the model's parameters (\emph{parametric} memory) and external information (\emph{non-parametric} memory), as shown by \newcite{lewis_2020}. An example of this is the \textsc{Atlas} model \cite{izacard2023atlas}, which connects a language model to an external source of information and allows \textsc{Atlas} to handle tasks that need up-to-date or specialized knowledge.

\begin{table}[!t]
\centering
\small
\begin{tabular}{>{\raggedright\arraybackslash}m{4.5cm} >{\centering\arraybackslash}m{1.5cm} >{\centering\arraybackslash}m{0.5cm}}
\toprule
\textbf{Context} & \textbf{Output (O)} & $\bm{P_{O}}$ \\
\midrule
In 1634, Stockholm became the official capital of Sweden. & Stockholm & 0.84 \\
In 1634, Milan became the official capital of Sweden. & Milan & 0.95 \\
Milan has been the capital since 1634. & Stockholm & 0.51 \\
Milan became well-known in Sweden since 1634. & Milan & 0.98 \\
Milan became well-known since 1634. & Stockholm & 0.67 \\
In 1634, Milan became the official capital of Italy. & Stockholm & 0.64 \\
\bottomrule
\end{tabular}
\caption{Model behavior with different contexts for the question \emph{What is the capital of Sweden?} The table shows the predicted outputs (\textbf{O}) and the corresponding probability $P_O$ assigned by \textsc{Atlas}. The first row represents the baseline context. When the counterfactual ``Milan'' is added, if the model answers ``Milan,'' this shows that the model relies on its non-parametric mechanism rather than its parametric memory.}
\label{tab:intro}
\end{table}

Despite the success of RAG systems in knowledge-intensive tasks, several aspects of these systems remain poorly explored in the research community. The most important mechanism RAG models apply is to retrieve relevant passages from which information can be extracted, similar to classical open-book question-answering systems  \cite{norlund2023generalization}. On the other hand, RAG systems must still produce sensible answers even when the retrieved context is less useful. In such cases, the RAG systems generate answers based on the knowledge stored in their parameters similar to pure language models. The model's use of these two fundamental mechanisms -- the non-parametric mechanism, where the model \emph{copies} the answer from a relevant retrieved context and the parametric mechanism of \emph{recalling} an answer from memorized knowledge -- leads to questions about how these two mechanisms interact. Which mechanism is more important, and how does the model determine which to rely on when given a context?

As shown in Table \ref{tab:intro}, the model's behavior varies depending on the context of the same question: \emph{What is the capital of Sweden?} The outputs show the duality discussed above: the model sometimes relies on parametric memory with high confidence, while at other times, it relies on non-parametric memory. This change in behavior highlights the complexity of the model's decision-making process within RAG models and offers an opportunity to learn more about how they work. 

In this study, we address two main research questions about how parametric and non-parametric memory interact within the \textsc{Atlas} model.

\begin{enumerate}\addtolength{\itemsep}{-0.5\baselineskip}
    \item Which aspect of the model representation impacts the output in copying mode? 
    \item What specific parts of the model trigger copying?
\end{enumerate}

Through two series of experiments, we aim to answer these two research questions and identify the factors that influence a model's dependence on its parametric memory versus its non-parametric memory. This understanding will help improve the way these models integrate and update information. The primary contributions of this paper are:

\begin{itemize}\addtolength{\itemsep}{-0.5\baselineskip}
    \item We examine how the ATLAS model makes decisions or simpler how it uses different types of memory. 
    \item We show when the model prefers to use one type of memory over the other, and how changes in context affect its decisions. 
    \item We also identify specific parts of the model that are crucial for copying and determining relevance.
\end{itemize}

\section{Method}
This study is inspired by previous work that applied causal mediation analysis to elucidate how language models process memorized knowledge stored in their parameters \cite{meng_2022}.
However, we revise and enhance our method to better suit our research questions about the interplay between parametric knowledge and contextual information. The following sections describe the theoretical framework we built upon, the experimental setup, and the details of the datasets and preprocessing.

\subsection{Background: Causal Mediation Analysis}
Our contribution follows the line of work that applies methods drawn from causal inference \cite{pearl2000causality} to analyze the behavior of models and their inner dynamics \cite{feder2022causal}. In particular, we apply \emph{causal mediation analysis} \cite{pearl2001effects} to investigate how specific parts of the model contribute to its overall behavior, following pioneering work by \newcite{vig_2020} who first applied mediation analysis for this purpose.

Causal mediation analysis can be applied when we want to disentangle the contribution to an overall effect of an individual component in a complex system. 

In this framework, a control variable $X$ affects an outcome $Y$, and we define the \emph{total effect} (TE) to quantify the impact of $X$ on $Y$.
\[
\text{TE} = Y(X\leftarrow 1) - Y(X\leftarrow 0)
\]
The notation $Y(X \leftarrow$1) corresponds to the \texttt{do} operator: the value of $Y$ when an intervention has been carried out that sets $X$ to 1.

However, the interaction between $X$ and $Y$ is complex because 
on the one hand there is a direct effect of $X$ on $Y$, and on the other hand also an indirect effect through a \emph{mediator} $M$. Mediation analysis introduces a framework to speak of the relative strengths of these different effects.

There are multiple ways to define the notion of direct and indirect effects \cite{pena2023alternative}. We follow previous work in model analysis by applying the framework of \emph{natural} effects by \newcite{pearl2001effects}. The \emph{natural indirect effect} (IE) is defined as follows:

\[
\text{IE} = Y(X\leftarrow 0, M(X\leftarrow 1)) - Y(X\leftarrow 0)
\]
The interpretation of this quantity is the expected change in the outcome variable if the mediator behaves as if $X$ were set to 1, while all other parts of the system behave as if $X$ were set to 0.

Causal mediation analysis provides a natural framework for investigating the behavior of complex systems such as neural NLP models \cite{vig_2020}. In this type of investigation, the mediator $M$ will typically correspond to an internal model representation, and it allows us to disentangle the contribution of this part from other parts of the model.

The control $X$ and the outcome $Y$ are defined in different ways depending on what research question is being investigated. \newcite{vig_2020} investigated gender bias ($Y$) using interventions on the text ($X$), while \newcite{meng_2022} investigated fact memorization by observing changes in next-token probabilities ($Y$) when running the model on clean or corrupted input embeddings ($X$).

In contrast to a causal inference situation based on observational data alone, computing the IE in model analysis is straightforward, since we can observe both outcomes ($X\leftarrow0$ and $X\leftarrow1$) by running the model with different inputs. To compute $Y(X\leftarrow0, M(X\leftarrow1))$, we first run the model with $X\leftarrow1$ to observe the intermediate representation $M$; we then run the model again with $X\leftarrow0$, while setting $M$ to the previously observed result. This is referred to by \newcite{meng_2022} as a \emph{corrupted with restoration} run.

\begin{table*}[t!]
\centering
\footnotesize
\begin{tabular}{>{\raggedright\arraybackslash}m{1.5cm} >{\raggedright\arraybackslash}m{5.35cm} >{\raggedright\arraybackslash}m{7.2cm} >{\centering\arraybackslash}m{0.3cm}}
\toprule
\textbf{Relations} & \textbf{Query Template} & \textbf{Context Template} & \textbf{\#} \\
\midrule
\textbf{capital} & What is the capital of \texttt{[subj]} ? & The capital of \texttt{[subj]} is \texttt{[obj]}. & 101 \\ 
\textbf{capital of} & What is \texttt{[subj]} the capital of ? & \texttt{[subj]} is the capital of \texttt{[obj]}. & 26 \\ 
\textbf{color} & What color is \texttt{[subj]} ? & The color of \texttt{[subj]} is \texttt{[obj]}. & 4 \\ 
\textbf{composer} & Who was the composer of \texttt{[subj]} ? & \texttt{[obj]} was the composer of the musical work \texttt{[subj]}. & 4 \\ 
\textbf{country} & In what country is \texttt{[subj]} ? & The \texttt{[subj]} is located in \texttt{[obj]}. & 101 \\ 
\textbf{father} & Who is the father of \texttt{[subj]} ? & \texttt{[obj]} is the father of \texttt{[subj]}. & 3 \\ 
\textbf{genre} & What genre is \texttt{[subj]}? & The work titled \texttt{[subj]} belongs to the \texttt{[obj]} genre. & 17 \\ 
\textbf{occupation} & What is \texttt{[subj]}'s occupation ? & The occupation of \texttt{[subj]} is \texttt{[obj]}. & 4 \\ 
\textbf{place of birth} & In what city was \texttt{[subj]} born ? & \texttt{[subj]} was born in the city of \texttt{[obj]}. & 13 \\ 
\textbf{religion} & What is the religion of \texttt{[subj]} ? & \texttt{[subj]} practices the \texttt{[obj]} religion. & 15 \\ 
\textbf{sport} & What sport does \texttt{[subj]} play ? & The \texttt{[subj]} team plays the sport of \texttt{[obj]}. & 20 \\ 
\textit{P17} & Which country is \texttt{[subj]} located in ? & \texttt{[subj]} is located in the country of \texttt{[obj]}. & 101 \\ 
\textit{P19} & Where was \texttt{[subj]} born ? & According to records, \texttt{[subj]} was born in \texttt{[obj]}. & 101 \\ 
\textit{P20} & Where did \texttt{[subj]} die ? & \texttt{[subj]} passed away in \texttt{[obj]}. & 101 \\ 
\textit{P36} & Where was \texttt{[subj]} born ? & According to records, the capital of \texttt{[subj]} is \texttt{[obj]}. & 83 \\ 
\textit{P69} & Where was \texttt{[subj]} educated ? & \texttt{[subj]} received their education at \texttt{[obj]}. & 16 \\ 
\textit{P106} & What kind of work does \texttt{[subj]} do ? & \texttt{[subj]} is employed as a \texttt{[obj]} according to structured data. & 14 \\ 
\textit{P127} & Who owns \texttt{[subj]} ? & \texttt{[subj]} is owned by \texttt{[obj]}. & 24 \\ 
\textit{P131} & Where is \texttt{[subj]} located ? & \texttt{[subj]} is located in \texttt{[obj]}. & 14 \\ 
\textit{P159} & Where is the headquarter of \texttt{[subj]} ? & The headquarters of \texttt{[subj]} is located in \texttt{[obj]}. & 101 \\ 
\textit{P175} & Who performed \texttt{[subj]} ? & \texttt{[obj]} performed the song \texttt{[subj]}. & 16 \\ 
\textit{P176} & Which company is \texttt{[subj]} produced by ? & The \texttt{[subj]} is produced by the company \texttt{[obj]}. & 66 \\ 
\textit{P276} & Where is \texttt{[subj]} located ? & The \texttt{[subj]} took place in \texttt{[obj]}. & 25 \\ 
\textit{P407} & Which language was \texttt{[subj]} written in ? & \texttt{[subj]} was written in the \texttt{[obj]} language. & 101 \\ 
\textit{P413} & What position does \texttt{[subj]} play ? & \texttt{[subj]} plays in the position of \texttt{[obj]}. & 14 \\ 
\textit{P495} & Which country was \texttt{[subj]} created in ? & \texttt{[subj]} was created in \texttt{[obj]}. & 101 \\ 
\textit{P740} & Where was \texttt{[subj]} founded ? & \texttt{[subj]} was founded in \texttt{[obj]}. & 60 \\ 
\bottomrule
\end{tabular}
\caption{Full list of the queries that were built using synthetic context templates derived from both datasets. \texttt{[subj]} and \texttt{[obj]} serve as placeholders for subject and object entities. Bold and italic styles are used to differentiate between the two datasets (PopQA and PEQ, respectively).} 
\label{tb:popqa_peq}
\end{table*}

\subsection{Experimental Design}
Throughout the paper, we investigate a series of questions relating to how much a RAG model favors an answer based on the retrieved context as opposed to the answer stored by its learned parameters. To disentangle these effects, we modify the context to replace the occurrence of the target entity by a \emph{counterfactual}: another entity of the same type. Intuitively, we can then investigate the research questions by considering the probabilities of the true answer in relation to the counterfactual. This idea is encoded in the outcome variable $Y$, which is defined as follows in all experiments:
\[
Y = \log \frac{P(\text{counterfactual} | \text{context})}{P(\text{true answer} | \text{context})}
\]
The investigations in this paper are carried out through two series of experiments, where we define the control variable $X$ in different ways. We introduce the log transformation to make subtle contributions visible and for numerical stability.

\begin{figure*}[t]
    \centering
    \includegraphics[width=\textwidth, keepaspectratio]{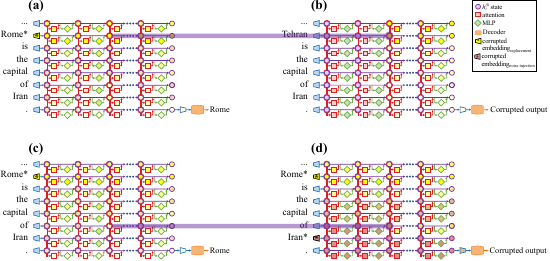}
    \caption{The first row represents the first experiment, while the second row represents the second experiment for the subjects in relation to the "restoration run". 
    (a) shows the representations with injected counterfactual embeddings for the query "What is the capital of Iran?" (b) depicts how restoration occurs at token $i$ and layer $l$.  Moving on to the second experiment, which is very similar to \newcite{meng_2022}: (c) shows the representations when we replace the object tokens ("Tehran") with a counterfactual ("Rome"). (d) demonstrates how restoration occurs after adding noise to the subject tokens ("Iran").  We have a similar implementation for the second experiment for relations.}
   \label{fig:experiments}
\end{figure*}

\paragraph{Experiment 1. What is the balance between parametric and non-parametric behavior?}

The first set of experiments investigates the degree to which the model copies from the context or relies on knowledge stored in its learned parameters, and which parts of the model are most impactful when the model is copying.
In these experiments, the control variable $X$ describes whether the context is unchanged ($X\leftarrow0$) or whether it has been modified so that occurrences of the true answer have been replaced with the counterfactual ($X\leftarrow1$), As illustrated in Figure~\ref{fig:experiments}.

The \emph{total effect} (TE) in this experiment measures how much we shift the model's output towards the counterfactual when modifying the retrieved context to replace the true answer with the counterfactual. Essentially, this quantifies the extent of the copying behavior: this quantity will be larger in cases where the model copies directly from the context. Conversely, if the prediction is based mostly on the stored parametric knowledge, the change in probabilities will be smaller.

By applying mediation analysis as described above, 
the \emph{indirect effect} (IE) 
quantifies the contribution of a selected intermediate representation $M$ to the overall copying behavior of the model.
Following previous work that applied mediation analysis to elucidate the behavior of complex models, we then carry out \emph{causal tracing} where we visualize the average of IE (AIE) for different tokens and layers to understand which parts are the most impactful.

\paragraph{Experiment 2. What makes the model decide to rely on the context?}
Intuitively, when presented with a retrieved context, the model makes a decision about whether the context is \emph{relevant} or not: whether it contains an answer that can be copied (relevance evaluation). The second set of experiments investigates how the model makes the decision about the relevance of the context. This decision is going to be affected by a multitude of factors;
in this work, we hypothesize that the presence of \emph{subject tokens} and \emph{relation tokens} in the context are important for this decision, and we leave the investigation of additional factors to future work.

In these experiments, we work with retrieved context documents where the true answers have been replaced with a counterfactual.
Similarly to the setup by \newcite{meng_2022}, the control variable $X$ in this case corresponds to whether the embeddings of the context subject tokens or relationship tokens have been affected by noise ($X\leftarrow0$) or are set to their original values ($X\leftarrow$1), As shown in Figure~\ref{fig:experiments}. In this second set of experiments, the average of TE (ATE) measures the shift towards the counterfactual when providing the model with uncorrupted subject or relation embeddings. The purpose of this measurement is to quantify the general impact of context subject tokens or context relation tokens on the model's decision to rely on the context or its learned parameters.
The AIE in this experiment shows how a selected representation $M$ contributes to this decision, and again we carry out causal tracing over the model to find the most impactful model components.

\subsection{Datasets}
We used two datasets for our study: PopQA \cite{popqa_ds} and PrincetonEntityQuestion (PEQ) \cite{peq_ds}. They both comprise entity-centric Question-Answer pairs (QAs) and include factual triples (subject, relation, object) associated with natural language queries \cite{nq_ds}. The PopQA prioritizes popular entities and includes a variety of relations, while the PEQ focuses on sentences found in Wikipedia that are rich in presence (more than 2,000 instances) and form straightforward questions. There are more than 10,000 factual questions in each dataset.

\subsection{The RAG System under Investigation: The \textsc{Atlas} Model}

Our investigation examines \textsc{Atlas} \cite{izacard2023atlas} as an example of a RAG model that integrates parametric and non-parametric components to utilize external data effectively. This integration, and the fact that \textsc{Atlas} is pre-trained jointly with a retriever, makes it well suited to studying how information is synthesized in response to a prompt. For this study, we use the version of \textsc{Atlas} that has been fine-tuned on Google's Natural Questions \cite{nq_ds}. We focus on only the language model (the sequential-to-sequential component) of \textsc{Atlas}, which uses the retrieved document related to the query to generate the answers.

\subsection{Data Preparation}
Actual retrieved documents have different quality of relations (everything in the context except the subject and object). Some retrieved documents may contain high-quality relational data that directly addresses or expands on the query, while others may introduce noise or irrelevant information. To be able to isolate the effects of relations, have fair comparison across data points, and create a consistent experimental setup, we consider the retrieved document for each query as a controlled template (a synthetic context), as shown in Table~\ref{tb:popqa_peq}. In Appendix~\ref{app:exp_real_context}, we show that the results on actual documents extracted using the built-in retriever in \textsc{Atlas} are similar.

Following \newcite{meng_2022}, to ensure that the model's parametric knowledge represents the answer, we retain only those samples for which the model generates the correct answer with and without their context. The correct answer may be a substring of the actual answer, such as \emph{Zaragoza} being a correct answer for \emph{Zaragoza, Spain} or \emph{Zaragoza city}. We removed relations where we had just a few data points after filtering. Table~\ref{tb:popqa_peq} shows the complete set of relations used in the experiments.

\subsection{Path Specific Effects (PSE): Implementation Details}
In our previous setup, we conducted experiments to investigate the impact of individual tokens at each layer concerning copying behavior and context relevance. However, we did not explore the separate effects of each module (\emph{MLP} and \emph{Attention}) to understand their individual contributions. To this end, we utilize the other experiment introduced by \newcite{meng_2022} as PSE. We start by collecting the embeddings of each \emph{MLP} and \emph{Attention} module with corrupted input before restoration as the zero states (the baseline condition with corrupted input). Then, to isolate the effect of each module during the restoration, we replace the representation of the module at token $i$ and layer $l$ with the one we have in zero states. In simpler terms, if we want to investigate the effect of \emph{MLP}, we first store the embedding representation of \emph{MLP} for all tokens and layers. Then, when moving to restoration, for instance, we want to restore $Attention_{l}^{i}$ concerning token $i$ and layer $l$, we do that and then restore all the \emph{MLP} layers to the zero states that we already stored. Unlike PSE, in standard IE, we do not restore the zero states after restoration.

\subsection{Causal Tracing: Implementation Details}
For causal tracing, we averaged causal traces across a set of prompts for each template and over all the templates. In these experiments, we computed the AIE at three points of the transformer modules: the \emph{hidden states} (the output of a transformer block), the \emph{MLP}, and the \emph{Attention}.

We follow \newcite{meng_2022} and aggregate over tokens, and we extend this approach to consider the context in addition to the question. To generate the counterfactual contexts, we replaced the object tokens with another set of tokens appearing as the object in some other example in the same relation.

Following \newcite{meng_2022}, we implemented the \emph{corrupted-with-restoration run} by restoring the clean run's result in six consecutive layers in \emph{MLP} and \emph{Attention} modules. We will later conduct PSE experiments to investigate the special role of \emph{MLP} and \emph{Attention} modules when we compute the impact of \emph{hidden states}. Compared to what \newcite{meng_2022} introduced, we divided the token spaces into 11 divisions for all the experiments to compute the average effect. These divisions include question, beginning of context, first subject token, middle subject tokens, last subject token, context in between tokens, first object token, middle object tokens, last object token, rest of context tokens, and last token.

\section{Results and Discussion}

We discuss the interpretation of the ATE, AIE, and PSE (via causal tracing) for the two experiments. 

\begin{figure*}[t]
    \centering
    
    \begin{subfigure}[b]{0.32\textwidth}
        \centering
        \includegraphics[width=\textwidth]{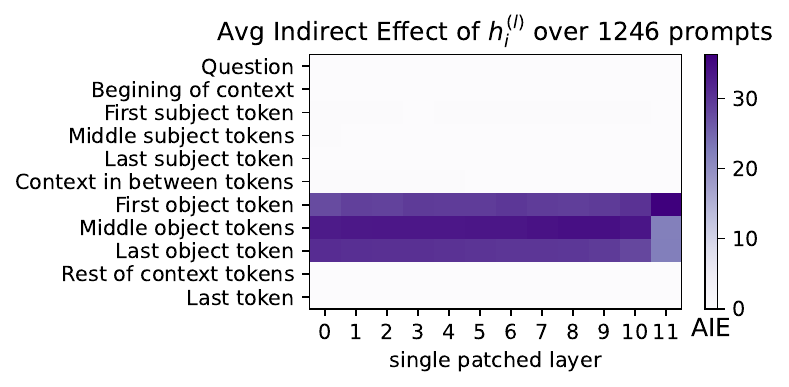}
        \caption{}
        \label{fig:aie_a_None}
    \end{subfigure}
    \hfill
    \begin{subfigure}[b]{0.32\textwidth}
        \centering
        \includegraphics[width=\textwidth]{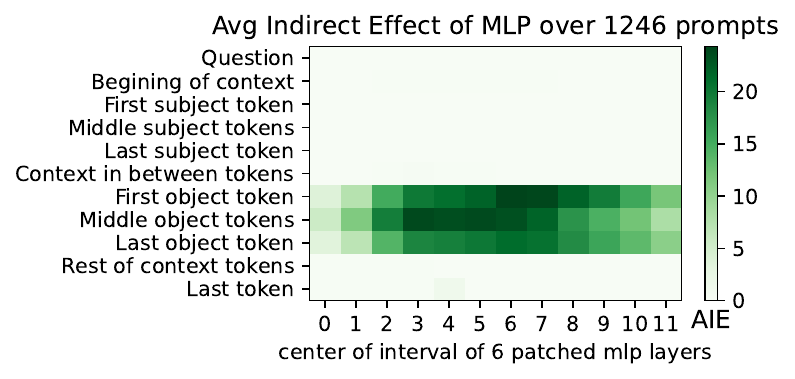}
        \caption{}
        \label{fig:aie_a_mlp}
    \end{subfigure}
    \hfill
    \begin{subfigure}[b]{0.32\textwidth}
        \centering
        \includegraphics[width=\textwidth]{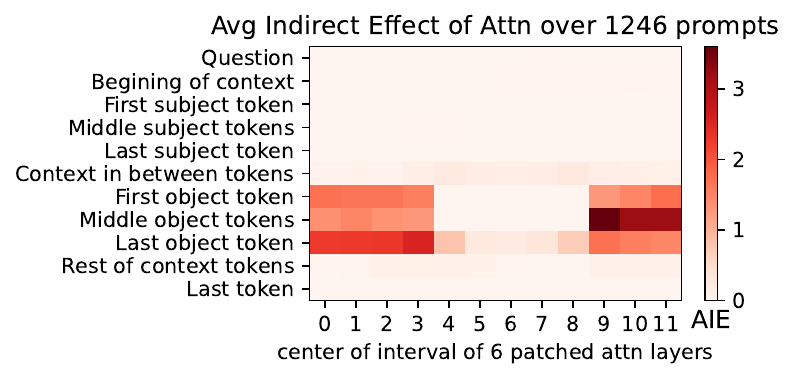}
        \caption{}
        \label{fig:aie_a_attn}
    \end{subfigure}

    \begin{subfigure}[b]{0.32\textwidth}
        \centering
        \includegraphics[width=\textwidth]{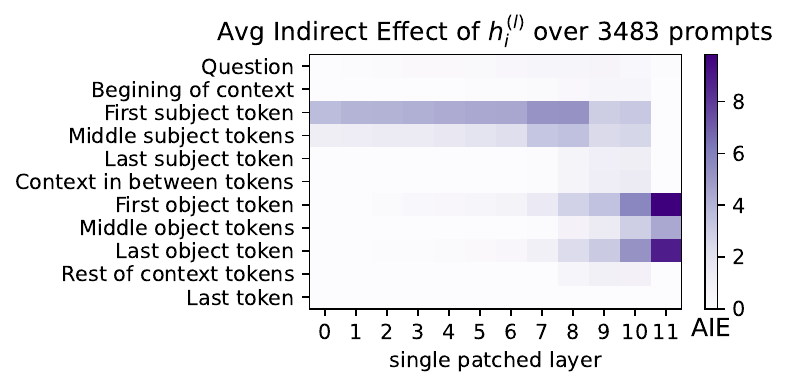}
        \caption{}
        \label{fig:aie_b_None}
    \end{subfigure}
    \hfill
    \begin{subfigure}[b]{0.32\textwidth}
        \centering
        \includegraphics[width=\textwidth]{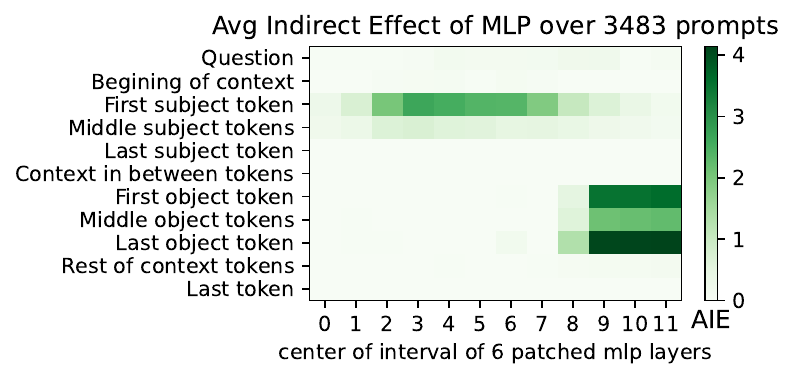}
        \caption{}
        \label{fig:aie_b_mlp}
    \end{subfigure}
    \hfill
    \begin{subfigure}[b]{0.32\textwidth}
        \centering
        \includegraphics[width=\textwidth]{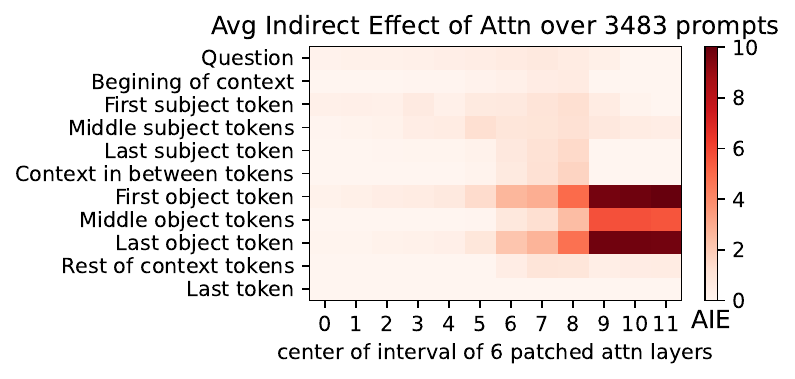}
        \caption{}
        \label{fig:aie_b_attn}
    \end{subfigure}

    \begin{subfigure}[b]{0.32\textwidth}
        \centering
        \includegraphics[width=\textwidth]{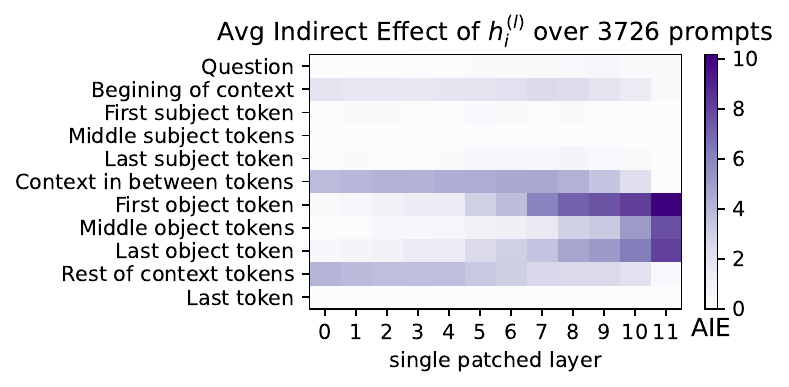}
        \caption{}
        \label{fig:aie_c_None}
    \end{subfigure}
    \hfill
    \begin{subfigure}[b]{0.32\textwidth}
        \centering
        \includegraphics[width=\textwidth]{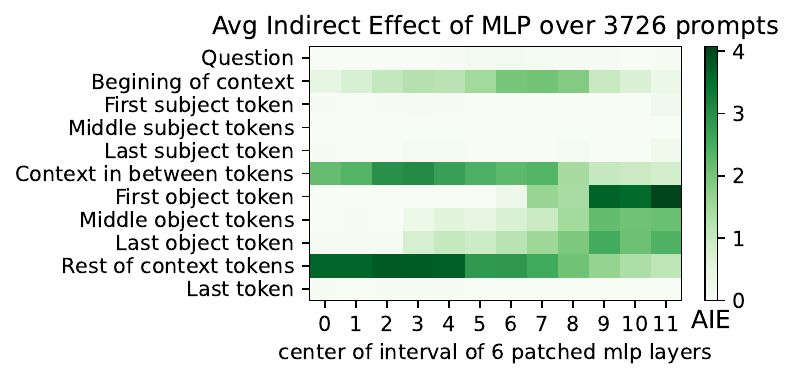}
        \caption{}
        \label{fig:aie_c_mlp}
    \end{subfigure}
    \hfill
    \begin{subfigure}[b]{0.32\textwidth}
        \centering
        \includegraphics[width=\textwidth]{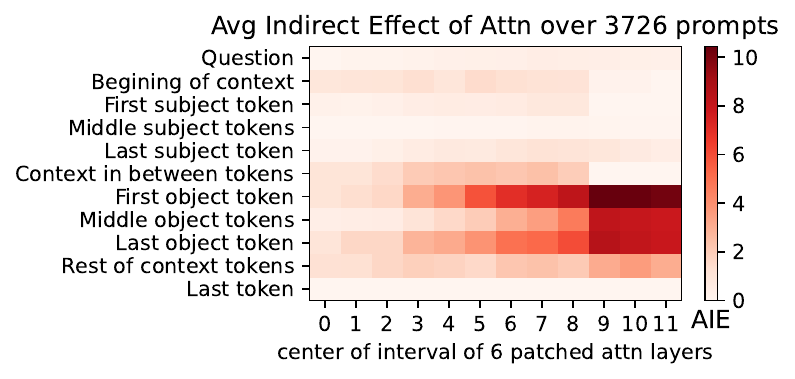}
        \caption{}
        \label{fig:aie_c_attn}
    \end{subfigure}
    
    \caption{The figures demonstrate the AIE results of the copying behavior in \textsc{Atlas} across different modules and layers. (a -- c) represent the AIEs of \emph{hidden states} (\(h^{(l)}\)), \emph{MLP}, and \emph{Attention} modules over the whole data points, which show that the object tokens are the dominant component in copying behavior. (d -- i) similarly, show the AIEs for the second experiment on subject and relations tokens respectively, highlighting the vital role of these two components in determining context relevancy.}
    \label{fig:aie_abc}
\end{figure*}

\paragraph{Experiment 1. Balance between parametric and non-parametric behavior}
The first set of experiments evaluated the model's responses when the object in the context was replaced with a counterfactual one. To understand the overall system's behavior, we categorized the results of experiments according to parametric and non-parametric. If the model consistently produces the correct answer despite counterfactual contexts, it should be classified as parametric;  otherwise, it shows non-parametric behavior.

\begin{figure}[!ht]
    \centering
    \includegraphics[width=\linewidth]{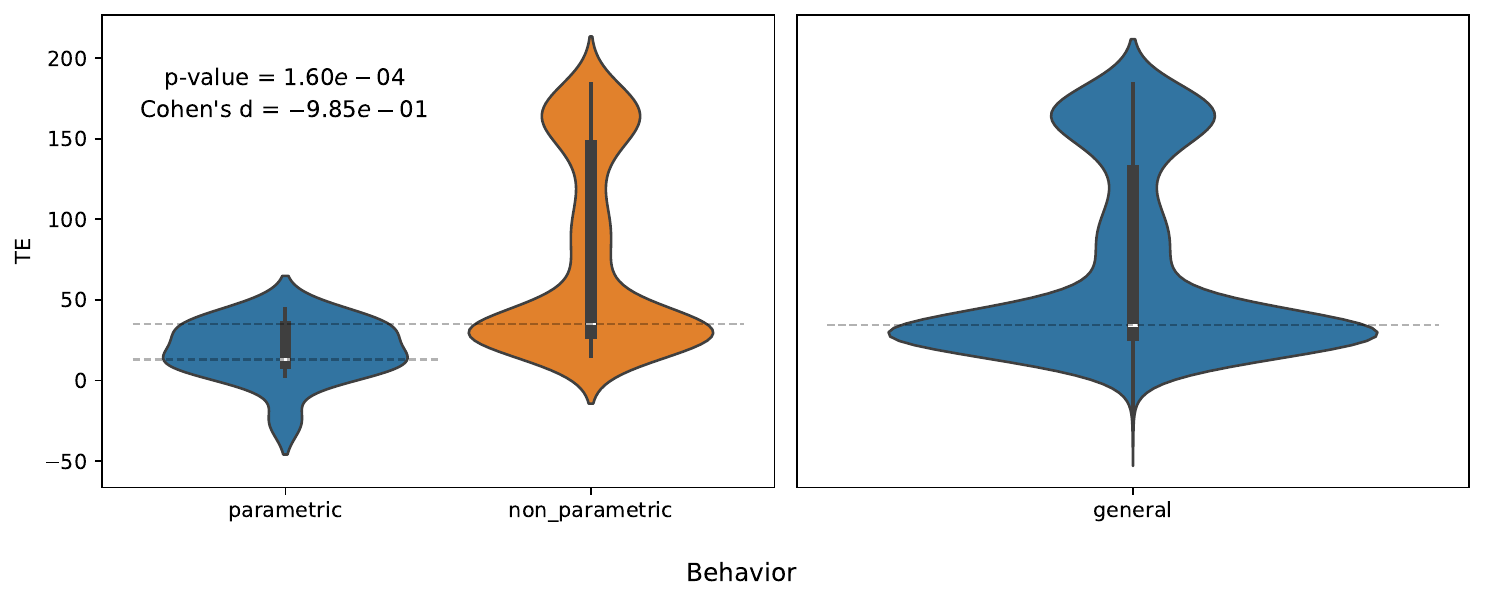}
    \caption{The left side illustrates the TE distribution across parametric and non-parametric behaviors, while the right side shows the overall distribution. The dominant distribution, represented in orange, indicates that the model's responses shift towards counterfactuals when the contexts are altered. Similarly, it reflects the model's general tendency to rely on the context to extract the answer (essentially, copying from the context).}
   \label{fig:te_ttest_a}
\end{figure}

A t-test \cite{student1908probable} and effect size analysis \cite{cohen1988statistical} (p-value=1.60e-4, Cohen's d=-0.9851), as shown in Figure~\ref{fig:te_ttest_a}, reveal statistically significant differences between these two categories, with the non-parametric subset showing much greater variability. This suggests that when the model engages in non-parametric behavior (copying from the context), it is susceptible to changes in the context. By analogy, it can be seen that the overall behavior of the model (as the general subset) is similar to the non-parametric subset, indicating a strong tendency of the model to copy from the contextual information.

\paragraph{Impactful tokens in copying situations}
The causal tracing results (Figure~\ref{fig:aie_a_None}--\ref{fig:aie_a_attn}) clearly show that object tokens are the most impactful when the model is in copying mode. The AIEs are close to zero for other token positions -- such as subject and relation tokens -- in the context. In these cases, the model performs a form of relevance evaluation. Once the model determines the context to be relevant for answering the query, it then copies relevant object tokens into the output.

\paragraph{Impactful components in copying situations}
The causal tracing (Figure~\ref{fig:aie_a_None}--\ref{fig:aie_a_attn}) across different model components (\emph{MLP} and \emph{Attention}) provides additional insights into how the model handles copying behavior. We observe that the object token representations flow directly through the model without being strongly affected by the surrounding context. Moreover, the \emph{MLP} in mid-layers plays a crucial role in translating representations from the encoder to the decoder. It needs to ensure that the copied object tokens can be passed into the decoder so that they can be generated as output. Since the encoder and decoder are in different latent spaces, the \emph{MLP} likely functions as the mechanism for this translation. This also explains that the \emph{Attention} shows lower AIEs in copying situations compared to the \emph{MLP}. \emph{Attention} may play a more supportive role, ensuring that the copied tokens stay coherent with the rest of the context.

\begin{figure*}[t]
    \centering
    \begin{subfigure}[b]{0.32\textwidth}
        \centering
        \includegraphics[width=0.95\textwidth]{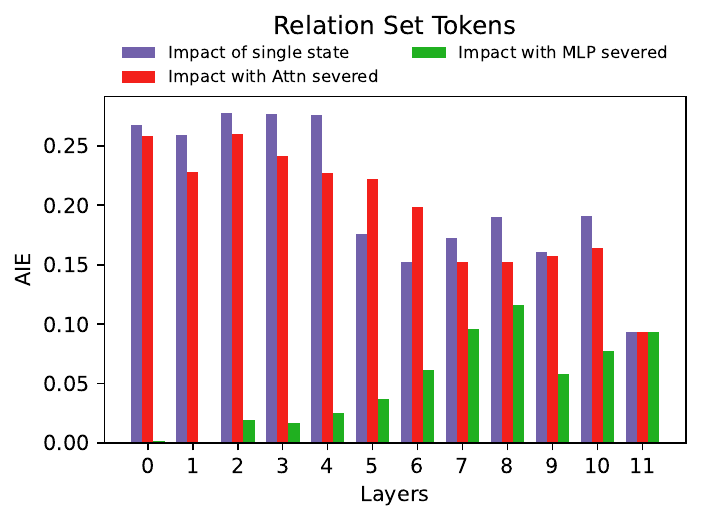}
        \caption{}
        \label{fig:impact_aie_a_relation}
    \end{subfigure}
    \hfill
    \begin{subfigure}[b]{0.32\textwidth}
        \centering
        \includegraphics[width=0.95\textwidth]{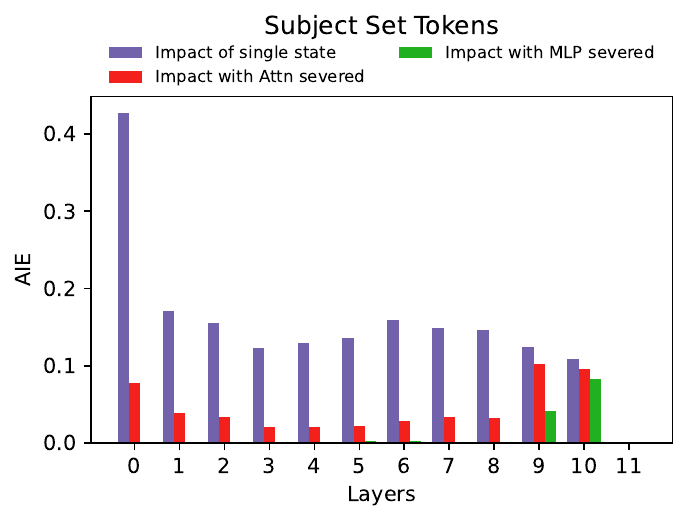}
        \caption{}
        \label{fig:impact_aie_a_subject}
    \end{subfigure}
    \hfill
    \begin{subfigure}[b]{0.32\textwidth}
        \centering
        \includegraphics[width=0.95\textwidth]{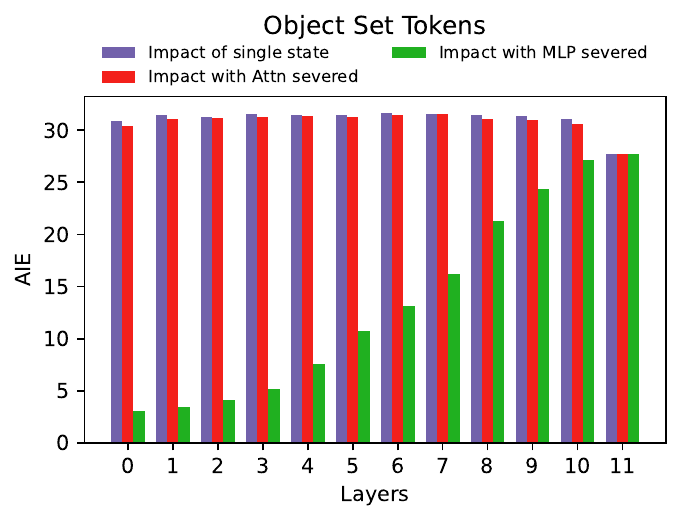}
        \caption{}
        \label{fig:impact_aie_a_object}
    \end{subfigure}

    \begin{subfigure}[b]{0.32\textwidth}
        \centering
        \includegraphics[width=0.95\textwidth]{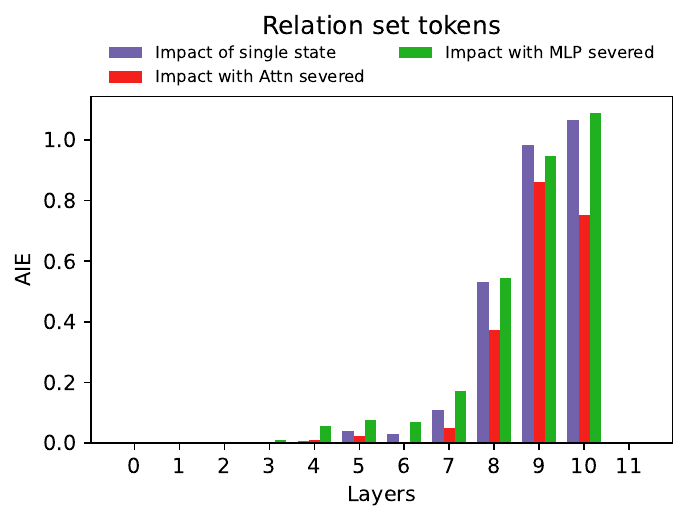}
        \caption{}
        \label{fig:impact_aie_b_relation}
    \end{subfigure}
    \hfill
    \begin{subfigure}[b]{0.32\textwidth}
        \centering
        \includegraphics[width=0.95\textwidth]{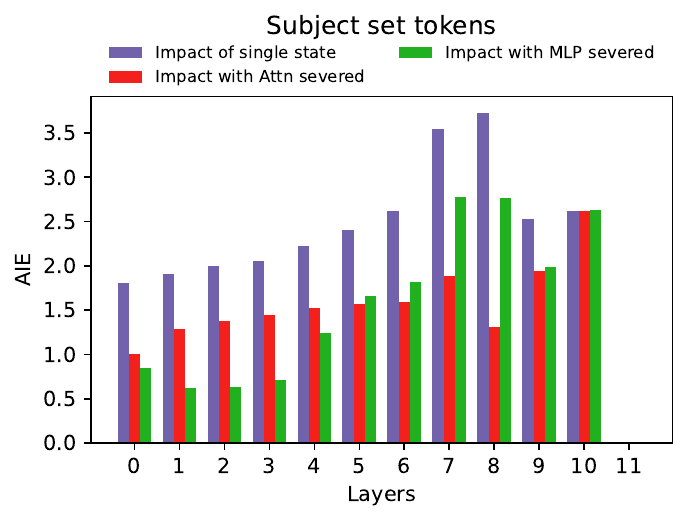}
        \caption{}
        \label{fig:impact_aie_b_subject}
    \end{subfigure}
    \hfill
    \begin{subfigure}[b]{0.32\textwidth}
        \centering
        \includegraphics[width=0.95\textwidth]{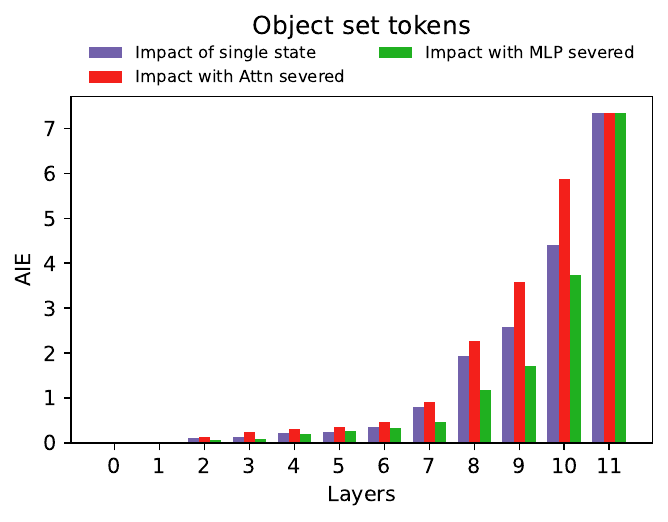}
        \caption{}
        \label{fig:impact_aie_b_object}
    \end{subfigure}

    \begin{subfigure}[b]{0.32\textwidth}
        \centering
        \includegraphics[width=0.95\textwidth]{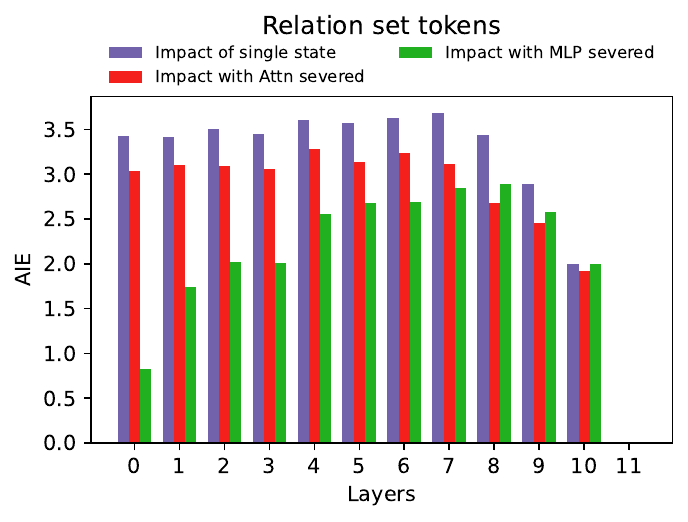}
        \caption{}
        \label{fig:impact_aie_c_relation}
    \end{subfigure}
    \hfill
    \begin{subfigure}[b]{0.32\textwidth}
        \centering
        \includegraphics[width=0.95\textwidth]{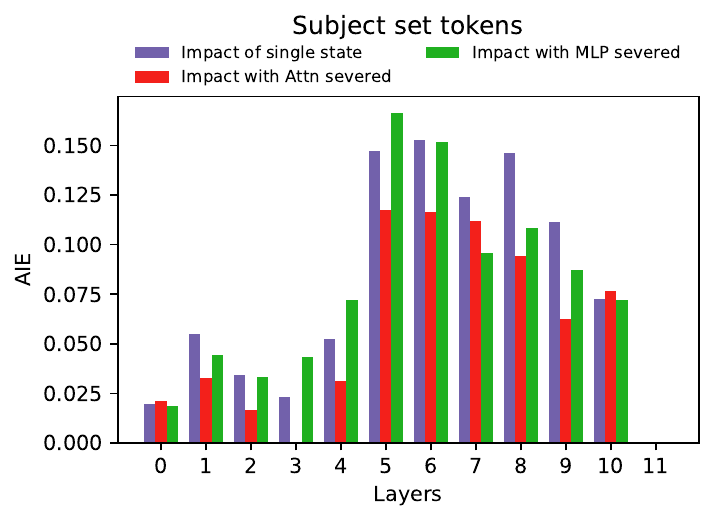}
        \caption{}
        \label{fig:impact_aie_c_subject}
    \end{subfigure}
    \hfill
    \begin{subfigure}[b]{0.32\textwidth}
        \centering
        \includegraphics[width=0.95\textwidth]{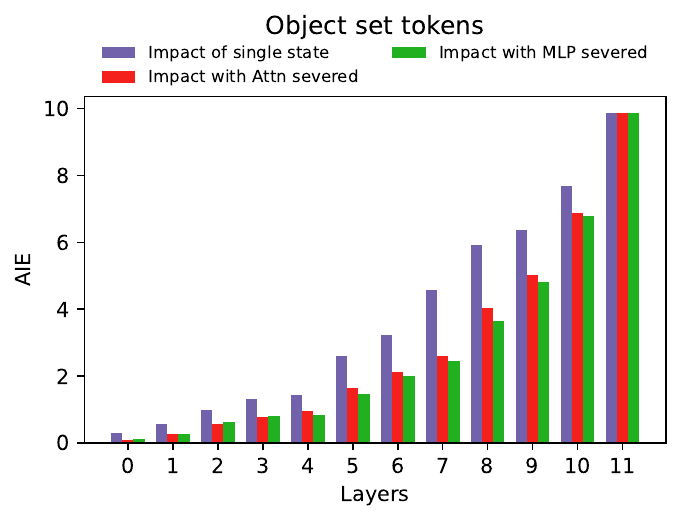}
        \caption{}
        \label{fig:impact_aie_c_object}
    \end{subfigure}
    
    \caption{These figures illustrate the impact of \emph{MLP} and \emph{Attention} on both earlier experiments. We consider the average impact over all the subject, object, and relation tokens as set tokens. (a--c) show the contribution of \emph{MLP} blocks from the early to the middle layers are key contributors to the model's ability to translate object token representations from the encoder space to the decoder while the \emph{Attention} plays a minor role in the later layers. (d -- i) depict the contribution of both model components from the early to the later layers, aligning with the processing of context relevance and the extraction of object tokens.}
    
    \label{fig:impact_aie_abc}
\end{figure*}

PSE analysis (Figure~\ref{fig:impact_aie_a_relation}--\ref{fig:impact_aie_a_object}) provides a better resolution of impactful components during copying. We have observed that when the \emph{MLP} is severed, the model's ability to rely on object tokens is substantially reduced, particularly in the mid-layers. This supports \emph{MLP} is responsible for translating the object tokens into a form that can be passed to the decoder. The lower impact of \emph{Attention} suggests that it is less involved in this process. Interestingly, the \emph{MLP} also shows a similar effect for relation tokens, which may indicate that both object and relation tokens go through similar representation transformations before being passed to the decoder.

\begin{figure}[!ht]
    \centering
    \includegraphics[width=\linewidth]{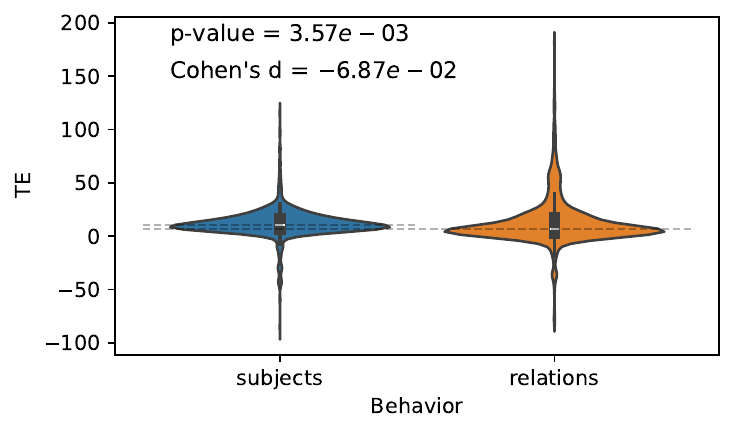}
    \caption{This plot shows the TE distribution across subjects and relation tokens. }
   \label{fig:te_ttest_abc}
\end{figure}

\paragraph{Experiment 2. Impact of rest of the context on the relevance mechanism}
In this experiment, we investigated the questions related to the earlier observations about how the model views context as relevant. We performed separate analyses on the two categories of tokens (subject and relation tokens) to determine their relative importance in relevance evaluation. The results, shown in Figure~\ref{fig:te_ttest_abc}, indicate that while there is a statistically significant difference between the effects of subject and relation tokens (p-value=3.57e-3), the effect size is quite small (Cohen's d=-6.87e-2), meaning that both types of tokens contribute similarly to the relevance process. Interestingly, the ATE distribution for subjects shows a slightly larger spread than relation tokens, suggesting that subjects may have a marginally greater influence on relevance.

\paragraph{Model layers in relation to  context relevance}
We observe an interesting pattern by looking at AIE values (Figure~\ref{fig:aie_b_None}--\ref{fig:aie_c_attn}) on how the model evaluates relevance. Low AIE values for object tokens in the early layers show that the model mainly focuses on subject and relation tokens in these layers. It is as if the model is first trying to determine the relevance of the context. As the processing moves to the middle and later layers, the higher AIE values in the last layer show that the focus gradually shifts toward object tokens. The \emph{MLP} and \emph{Attention} are key in transitioning from relevance evaluation to object extracting. The \emph{MLP} processes subject and relation tokens in the early layers, contributing to the context relevancy. Meanwhile, \emph{Attention} integrates these tokens by focusing on the entire context, ensuring the model maintains coherence across the context in this process. In the later layers, the \emph{MLP}'s role expands to help transform object token representations for the object extraction step, indicating the dual role of the \emph{MLP}.

The PSE results show that in the early layers, both the \emph{MLP} and \emph{Attention} work closely with subject and relation tokens (Figure~\ref{fig:impact_aie_b_relation}--\ref{fig:impact_aie_c_object}). Put simply, the model evaluates these tokens together to figure out if the context is relevant. As we move into the later layer, the \emph{Attention} becomes increasingly engaged with the object tokens. The change in object tokens shows that the \emph{MLP}, which was initially focused on subject and relation tokens, now works with \emph{Attention} to accurately extract the object tokens as the final answer.

\section{Related Work}
Recent research has increasingly focused on studying how Language Models (LMs) behave in specific situations. Although this field is still developing, several important studies have emerged. For example, \newcite{clasheval_2024} look into how large language models (LLMs) handle retrieved information that contains incorrect information. Their work involves in building small to large incorrect information to observe how models react. While parallel studies have been conducted on memory interplay in RAG models \cite{wadhwa2024ragsrichparametersprobing}, there is still a need to investigate the behavior of these models further. \newcite{roberts2020much} show that LLMs can answer questions by using the knowledge they learned during pre-training without needing external information. Similarly, \newcite{chen2022rich} study how LMs remember facts when they find conflicting information from different sources. \newcite{decao2021editing} suggest methods to update factual knowledge in models without needing much re-training. Their method uses a hyper-network to update the knowledge stored in the model's parameters.

In a related study, \newcite{longpre2021entity} explore how conflicts between contextual and parametric knowledge affect question-answering systems. Building on this idea, \newcite{wang2023resolving} develop a way to test how well models can find and resolve conflicts in contextual information. Their results show that while models can spot conflicting information, they often struggle to identify the exact parts that are in conflict and have difficulty producing clear responses that address all the different pieces of information.

\section{Conclusion}
The study offers valuable insights into the inner workings of the \textsc{Atlas} model -- a fine-tuned RAG model -- and how this model processes information from external sources (non-parametric memory) and learned parameters (parametric memory) in different queries. To clarify this phenomenon, we conducted two sets of experiments to understand how the model decides to choose copying from external sources over recalling from learned parameters. 

In the first experiment, we replaced the object tokens with counterfactuals to separate the effects of context from parametric knowledge. The results revealed that the model relies heavily on the context and tends to copy from it. The second experiment revealed how the model decides to rely on non-parametric knowledge and which mechanism causes the model to choose to copy over recall. Our results suggested that the model performs a relevance evaluation to ensure that the context is useful. If that is the case, it shifts focus towards identifying entities that might be replicated as the answer (the object tokens). The subject and relation tokens are essential here in determining context relevance.

Furthermore, both experiments' results indicated that the \emph{MLP} in the early to middle layers is involved in contextualizing the relevance of the tokens, while \emph{Attention} in the later layers helps the model focus on integrating this information to extract object tokens as the final answer. These findings explain how these models behave and open doors to controlling this behavior in future research.
 
\section*{Acknowledgements}
We express our gratitude to the anonymous reviewers for their insightful feedback and to Milad Malekipirbazari and Mohammad M. Ahmadpanah for their valuable input. This research was conducted 
in the project \emph{Representation Learning for Conversational AI}
under the Wallenberg AI, Autonomous Systems, and Software Program (WASP), funded by the Knut and Alice Wallenberg Foundation. We also appreciate the Swedish National Infrastructure for Computing (SNIC) for providing computational resources through grant agreement no. 2023/22-1025. 
We also would like to recognize our use of Grammarly\footnote{\href{https://www.grammarly.com/}{grammarly.com}} for paraphrasing and providing grammar suggestions.

\section*{Limitations}
In this section, we discuss the limitations of our study for future research.

\emph{Dataset Specificity}: We conducted our experiments using specific datasets (PopQA and PEQ) -- parametric and non-parametric memory behavior may differ between different datasets -- which may limit our findings' generalizability.

\emph{Context Manipulation}: In our study, we have used counterfactuals, which might not fully capture the actual situation where the context is noisy or ambiguous.

\emph{Model Generalization}: Although the \textsc{Atlas} model performed well in our experiments, Its versatility to other RAG models remains unknown. \textsc{Atlas} has been exposed to contexts of varying quality during training and must develop ways to adapt to poor-quality contexts, relying on parametric knowledge when necessary. It is unclear whether similar behavior can be observed in \emph{in-context} RAG implementations based on off-the-shelf LLMs that have not been trained in a RAG setup \cite{ram2023incontext}, or whether such models behave differently when deciding whether a context passage is relevant or not.

\emph{Temporal Relevance}: There is a potential limitation when dealing with outdated or rapidly changing information based on non-parametric memory. It is necessary to understand how the model adjusts to the temporal changes in knowledge and how effectively it can choose parametric versus non-parametric memory when these changes occur.

\section*{Ethics Statement}

Our work is in the area of analysis of existing models and we do not release any new model as part of this project, so we see no obvious ways to abuse the results presented here.

\bibliography{anthology,custom}

\begin{thebibliography}{25}
\expandafter\ifx\csname natexlab\endcsname\relax\def\natexlab#1{#1}\fi

\bibitem[{Chen et~al.(2022)Chen, Zhang, and Choi}]{chen2022rich}
Hung-Ting Chen, Michael Zhang, and Eunsol Choi. 2022.
\newblock \href {https://doi.org/10.18653/v1/2022.emnlp-main.146} {Rich knowledge sources bring complex knowledge conflicts: Recalibrating models to reflect conflicting evidence}.
\newblock In \emph{Proceedings of the 2022 Conference on Empirical Methods in Natural Language Processing}, pages 2292--2307, Abu Dhabi, United Arab Emirates. Association for Computational Linguistics.

\bibitem[{Cohen(1988)}]{cohen1988statistical}
Jacob Cohen. 1988.
\newblock \href {https://www.taylorfrancis.com/books/mono/10.4324/9780203771587/statistical-power-analysis-behavioral-sciences-jacob-cohen} {\emph{Statistical power analysis for the behavioral sciences}}, 2nd edition.
\newblock Routledge, New York.

\bibitem[{De~Cao et~al.(2021)De~Cao, Aziz, and Titov}]{decao2021editing}
Nicola De~Cao, Wilker Aziz, and Ivan Titov. 2021.
\newblock \href {https://doi.org/10.18653/v1/2021.emnlp-main.522} {Editing factual knowledge in language models}.
\newblock In \emph{Proceedings of the 2021 Conference on Empirical Methods in Natural Language Processing}, pages 6491--6506, Online and Punta Cana, Dominican Republic. Association for Computational Linguistics.

\bibitem[{Feder et~al.(2022)Feder, Keith, Manzoor, Pryzant, Sridhar, Wood-Doughty, Eisenstein, Grimmer, Reichart, Roberts, Stewart, Veitch, and Yang}]{feder2022causal}
Amir Feder, Katherine~A. Keith, Emaad Manzoor, Reid Pryzant, Dhanya Sridhar, Zach Wood-Doughty, Jacob Eisenstein, Justin Grimmer, Roi Reichart, Margaret~E. Roberts, Brandon~M. Stewart, Victor Veitch, and Diyi Yang. 2022.
\newblock \href {https://doi.org/10.1162/tacl_a_00511} {Causal inference in natural language processing: Estimation, prediction, interpretation and beyond}.
\newblock \emph{Transactions of the Association for Computational Linguistics}, 10:1138--1158.

\bibitem[{Hagstr{\"o}m et~al.(2023)Hagstr{\"o}m, Saynova, Norlund, Johansson, and Johansson}]{hagstrom2023effect}
Lovisa Hagstr{\"o}m, Denitsa Saynova, Tobias Norlund, Moa Johansson, and Richard Johansson. 2023.
\newblock \href {https://doi.org/10.18653/v1/2023.emnlp-main.332} {The effect of scaling, retrieval augmentation and form on the factual consistency of language models}.
\newblock In \emph{Proceedings of the 2023 Conference on Empirical Methods in Natural Language Processing}, pages 5457--5476, Singapore. Association for Computational Linguistics.

\bibitem[{Izacard et~al.(2023)Izacard, Lewis, Lomeli, Hosseini, Petroni, Schick, Dwivedi-Yu, Joulin, Riedel, and Grave}]{izacard2023atlas}
Gautier Izacard, Patrick Lewis, Maria Lomeli, Lucas Hosseini, Fabio Petroni, Timo Schick, Jane Dwivedi-Yu, Armand Joulin, Sebastian Riedel, and Edouard Grave. 2023.
\newblock \href {http://jmlr.org/papers/v24/23-0037.html} {\textsc{Atlas}: Few-shot learning with retrieval augmented language models}.
\newblock \emph{Journal of Machine Learning Research}, 24(251):1--43.

\bibitem[{Kandpal et~al.(2023)Kandpal, Deng, Roberts, Wallace, and Raffel}]{kandpal2023large}
Nikhil Kandpal, Haikang Deng, Adam Roberts, Eric Wallace, and Colin Raffel. 2023.
\newblock \href {https://proceedings.mlr.press/v202/kandpal23a/kandpal23a.pdf} {Large language models struggle to learn long-tail knowledge}.
\newblock In \emph{Proceedings of the 40th International Conference on Machine Learning}, ICML'23. JMLR.org.

\bibitem[{Kwiatkowski et~al.(2019)Kwiatkowski, Palomaki, Redfield, Collins, Parikh, Alberti, Epstein, Polosukhin, Devlin, Lee, Toutanova, Jones, Kelcey, Chang, Dai, Uszkoreit, Le, and Petrov}]{nq_ds}
Tom Kwiatkowski, Jennimaria Palomaki, Olivia Redfield, Michael Collins, Ankur Parikh, Chris Alberti, Danielle Epstein, Illia Polosukhin, Jacob Devlin, Kenton Lee, Kristina Toutanova, Llion Jones, Matthew Kelcey, Ming-Wei Chang, Andrew~M. Dai, Jakob Uszkoreit, Quoc Le, and Slav Petrov. 2019.
\newblock \href {https://doi.org/10.1162/tacl_a_00276} {Natural questions: A benchmark for question answering research}.
\newblock \emph{Transactions of the Association for Computational Linguistics}, 7:452--466.

\bibitem[{Lewis et~al.(2020)Lewis, Perez, Piktus, Petroni, Karpukhin, Goyal, Kuttler, Lewis, tau Yih, Rockt{\"a}schel, Riedel, and Kiela}]{lewis_2020}
Patrick Lewis, Ethan Perez, Aleksandara Piktus, Fabio Petroni, Vladimir Karpukhin, Naman Goyal, Heinrich Kuttler, Mike Lewis, Wen tau Yih, Tim Rockt{\"a}schel, Sebastian Riedel, and Douwe Kiela. 2020.
\newblock \href {https://api.semanticscholar.org/CorpusID:218869575} {Retrieval-augmented generation for knowledge-intensive {NLP} tasks}.
\newblock \emph{ArXiv}, abs/2005.11401.

\bibitem[{Longpre et~al.(2021)Longpre, Perisetla, Chen, Ramesh, DuBois, and Singh}]{longpre2021entity}
Shayne Longpre, Kartik Perisetla, Anthony Chen, Nikhil Ramesh, Chris DuBois, and Sameer Singh. 2021.
\newblock \href {https://doi.org/10.18653/v1/2021.emnlp-main.565} {Entity-based knowledge conflicts in question answering}.
\newblock In \emph{Proceedings of the 2021 Conference on Empirical Methods in Natural Language Processing}, pages 7052--7063, Online and Punta Cana, Dominican Republic. Association for Computational Linguistics.

\bibitem[{Mallen et~al.(2023)Mallen, Asai, Zhong, Das, Khashabi, and Hajishirzi}]{popqa_ds}
Alex Mallen, Akari Asai, Victor Zhong, Rajarshi Das, Daniel Khashabi, and Hannaneh Hajishirzi. 2023.
\newblock \href {https://doi.org/10.18653/v1/2023.acl-long.546} {When not to trust language models: Investigating effectiveness of parametric and non-parametric memories}.
\newblock In \emph{Proceedings of the 61st Annual Meeting of the Association for Computational Linguistics (Volume 1: Long Papers)}, pages 9802--9822, Toronto, Canada. Association for Computational Linguistics.

\bibitem[{Meng et~al.(2022)Meng, Bau, Andonian, and Belinkov}]{meng_2022}
Kevin Meng, David Bau, Alex Andonian, and Yonatan Belinkov. 2022.
\newblock \href {https://proceedings.neurips.cc/paper_files/paper/2022/file/6f1d43d5a82a37e89b0665b33bf3a182-Paper-Conference.pdf} {Locating and editing factual associations in {GPT}}.
\newblock In \emph{Advances in Neural Information Processing Systems}, volume~35, pages 17359--17372. Curran Associates, Inc.

\bibitem[{Norlund et~al.(2023)Norlund, Doostmohammadi, Johansson, and Kuhlmann}]{norlund2023generalization}
Tobias Norlund, Ehsan Doostmohammadi, Richard Johansson, and Marco Kuhlmann. 2023.
\newblock \href {https://doi.org/10.18653/v1/2023.findings-eacl.109} {On the generalization ability of retrieval-enhanced transformers}.
\newblock In \emph{Findings of the Association for Computational Linguistics: EACL 2023}, pages 1485--1493, Dubrovnik, Croatia. Association for Computational Linguistics.

\bibitem[{Pearl(2000)}]{pearl2000causality}
Judea Pearl. 2000.
\newblock \href {https://www.cambridge.org/core/books/causality/B0046844FAE10CBF274D4ACBDAEB5F5B} {\emph{Causality: Models, reasoning, and inference}}.
\newblock Cambridge University Press.

\bibitem[{Pearl(2001)}]{pearl2001effects}
Judea Pearl. 2001.
\newblock \href {https://dl.acm.org/doi/10.5555/2074022.2074073} {Direct and indirect effects}.
\newblock In \emph{Proceedings of the Seventeenth Conference on Uncertainty in Artificial Intelligence}, UAI'01, page 411–420, San Francisco, USA. Morgan Kaufmann Publishers Inc.

\bibitem[{Petroni et~al.(2019)Petroni, Rockt{\"a}schel, Riedel, Lewis, Bakhtin, Wu, and Miller}]{petroni2019language}
Fabio Petroni, Tim Rockt{\"a}schel, Sebastian Riedel, Patrick Lewis, Anton Bakhtin, Yuxiang Wu, and Alexander Miller. 2019.
\newblock \href {https://doi.org/10.18653/v1/D19-1250} {Language models as knowledge bases?}
\newblock In \emph{Proceedings of the 2019 Conference on Empirical Methods in Natural Language Processing and the 9th International Joint Conference on Natural Language Processing (EMNLP-IJCNLP)}, pages 2463--2473, Hong Kong, China. Association for Computational Linguistics.

\bibitem[{Peña(2023)}]{pena2023alternative}
José~M. Peña. 2023.
\newblock \href {https://arxiv.org/pdf/2306.01292} {Alternative measures of direct and indirect effects}.
\newblock \emph{ArXiv}, abs/2306.01292.

\bibitem[{Ram et~al.(2023)Ram, Levine, Dalmedigos, Muhlgay, Shashua, Leyton-Brown, and Shoham}]{ram2023incontext}
Ori Ram, Yoav Levine, Itay Dalmedigos, Dor Muhlgay, Amnon Shashua, Kevin Leyton-Brown, and Yoav Shoham. 2023.
\newblock \href {https://doi.org/10.1162/tacl_a_00605} {In-context retrieval-augmented language models}.
\newblock \emph{Transactions of the Association for Computational Linguistics}, 11:1316--1331.

\bibitem[{Roberts et~al.(2020)Roberts, Raffel, and Shazeer}]{roberts2020much}
Adam Roberts, Colin Raffel, and Noam Shazeer. 2020.
\newblock \href {https://doi.org/10.18653/v1/2020.emnlp-main.437} {How much knowledge can you pack into the parameters of a language model?}
\newblock In \emph{Proceedings of the 2020 Conference on Empirical Methods in Natural Language Processing (EMNLP)}, pages 5418--5426, Online. Association for Computational Linguistics.

\bibitem[{Sciavolino et~al.(2021)Sciavolino, Zhong, Lee, and Chen}]{peq_ds}
Christopher Sciavolino, Zexuan Zhong, Jinhyuk Lee, and Danqi Chen. 2021.
\newblock \href {https://doi.org/10.18653/v1/2021.emnlp-main.496} {Simple entity-centric questions challenge dense retrievers}.
\newblock In \emph{Proceedings of the 2021 Conference on Empirical Methods in Natural Language Processing}, pages 6138--6148, Online and Punta Cana, Dominican Republic. Association for Computational Linguistics.

\bibitem[{Student(1908)}]{student1908probable}
Student. 1908.
\newblock \href {https://www.jstor.org/stable/2331554} {The probable error of a mean}.
\newblock \emph{Biometrika}, pages 1--25.

\bibitem[{Vig et~al.(2020)Vig, Gehrmann, Belinkov, Qian, Nevo, Singer, and Shieber}]{vig_2020}
Jesse Vig, Sebastian Gehrmann, Yonatan Belinkov, Sharon Qian, Daniel Nevo, Yaron Singer, and Stuart Shieber. 2020.
\newblock \href {https://proceedings.neurips.cc/paper_files/paper/2020/file/92650b2e92217715fe312e6fa7b90d82-Paper.pdf} {Investigating gender bias in language models using causal mediation analysis}.
\newblock In \emph{Advances in Neural Information Processing Systems}, volume~33, pages 12388--12401. Curran Associates, Inc.

\bibitem[{Wadhwa et~al.(2024)Wadhwa, Seetharaman, Aggarwal, Ghosh, Basu, Srinivasan, Zhao, Chaudhari, and Aghazadeh}]{wadhwa2024ragsrichparametersprobing}
Hitesh Wadhwa, Rahul Seetharaman, Somyaa Aggarwal, Reshmi Ghosh, Samyadeep Basu, Soundararajan Srinivasan, Wenlong Zhao, Shreyas Chaudhari, and Ehsan Aghazadeh. 2024.
\newblock \href {http://arxiv.org/abs/2406.12824} {From rags to rich parameters: Probing how language models utilize external knowledge over parametric information for factual queries}.

\bibitem[{Wang et~al.(2023)Wang, Feng, Wang, Shi, Balachandran, He, and Tsvetkov}]{wang2023resolving}
Yike Wang, Shangbin Feng, Heng Wang, Weijia Shi, Vidhisha Balachandran, Tianxing He, and Yulia Tsvetkov. 2023.
\newblock \href {https://arxiv.org/pdf/2310.00935} {Resolving knowledge conflicts in large language models}.
\newblock \emph{ArXiv}, abs/2310.00935.

\bibitem[{Wu et~al.(2024)Wu, Wu, and Zou}]{clasheval_2024}
Kevin Wu, Eric Wu, and James Zou. 2024.
\newblock \href {http://arxiv.org/abs/2404.10198} {Clasheval: Quantifying the tug-of-war between an llm's internal prior and external evidence}.

\end{thebibliography}

\clearpage
\newpage

\appendix
\onecolumn
\newpage
\section{Experiments on Real Contexts}
\label{app:exp_real_context}

We use the built-in retriever in \textsc{Atlas} to fetch 20 documents per query from a given dataset for the actual context section. We then filter these documents based on specific criteria:

\begin{enumerate}\addtolength{\itemsep}{-0.5\baselineskip}
\item We keep only documents with one subject and one object in their context, ensuring that the object does not appear in the question and one subject also appears in the query. The reason for this filter is to ensure that it is possible to output a correct answer given a context.
\item We apply a second filter to retain only those samples for which the model can generate the correct answer even without their retrieved documents. We then expand the datapoints by considering one document from the filtered set for each question that generated the correct answer. 
\end{enumerate}

\begin{figure*}[!ht]
    \centering
    
    \begin{subfigure}[b]{0.32\textwidth}
        \centering
        \includegraphics[width=\textwidth]{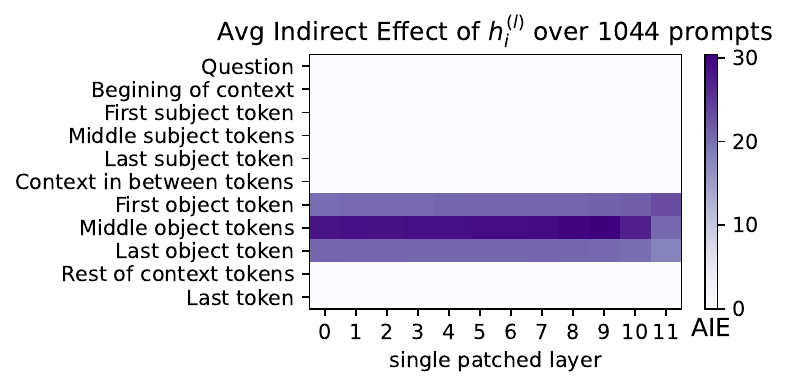}
        \caption{}
        \label{fig:aie_real_a_None}
    \end{subfigure}
    \hfill
    \begin{subfigure}[b]{0.32\textwidth}
        \centering
        \includegraphics[width=\textwidth]{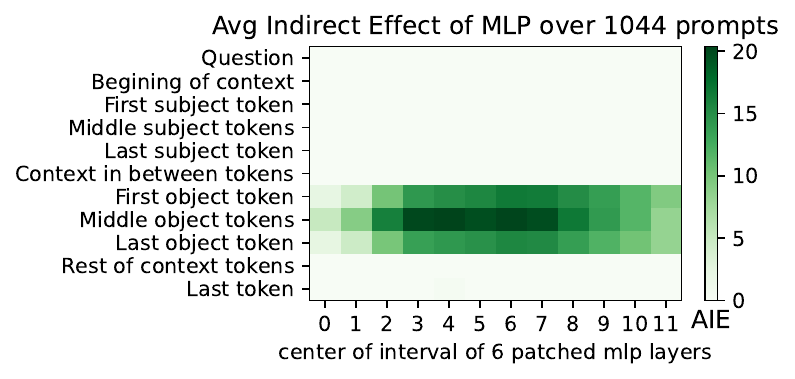}
        \caption{}
        \label{fig:aie_real_a_mlp}
    \end{subfigure}
    \hfill
    \begin{subfigure}[b]{0.32\textwidth}
        \centering
        \includegraphics[width=\textwidth]{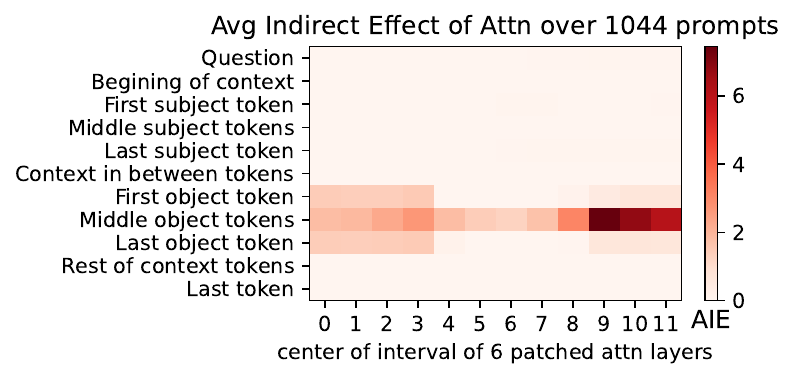}
        \caption{}
        \label{fig:aie_real_a_attn}
    \end{subfigure}

    \begin{subfigure}[b]{0.32\textwidth}
        \centering
        \includegraphics[width=\textwidth]{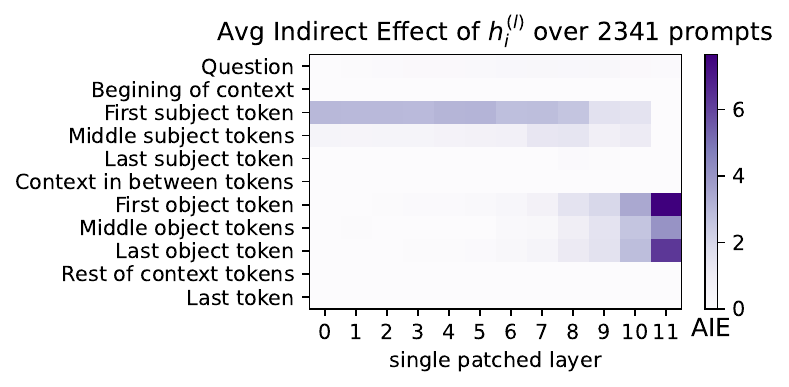}
        \caption{}
        \label{fig:aie_real_b_None}
    \end{subfigure}
    \hfill
    \begin{subfigure}[b]{0.32\textwidth}
        \centering
        \includegraphics[width=\textwidth]{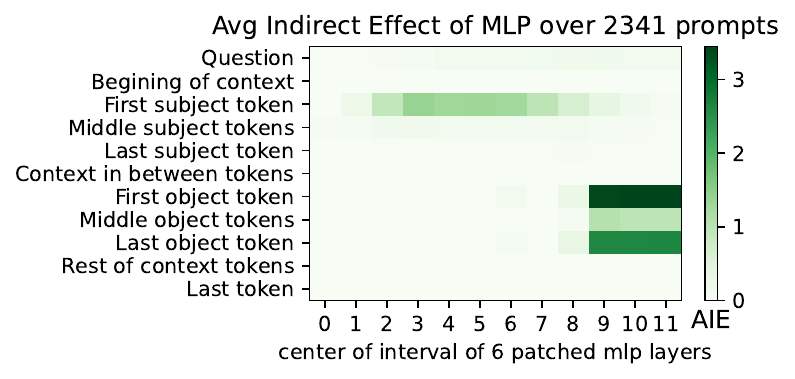}
        \caption{}
        \label{fig:aie_real_b_mlp}
    \end{subfigure}
    \hfill
    \begin{subfigure}[b]{0.32\textwidth}
        \centering
        \includegraphics[width=\textwidth]{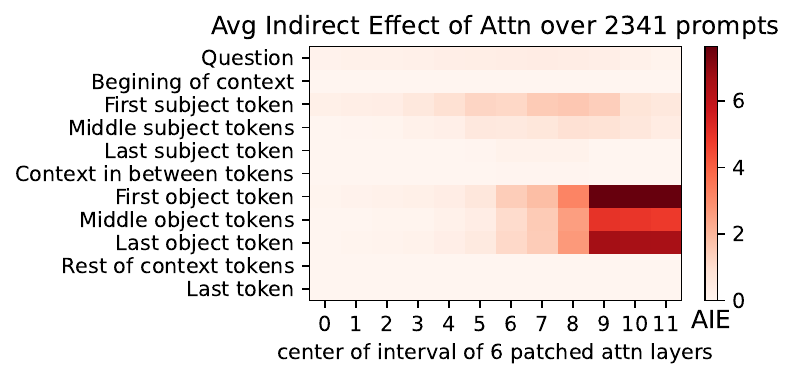}
        \caption{}
        \label{fig:aie_real_b_attn}
    \end{subfigure}

    \caption{The figures demonstrate the AIE results of the copying behavior in \textsc{Atlas} across different modules and layers for actual documents retrieved by the \textsc{Atlas} model's retriever. (a -- c) represent the AIEs of \emph{hidden states} (\(h^{(l)}\)), MLP, and \emph{Attention} modules over the whole prompts. (d -- f) similarly, show the AIEs for the second experiment on subject tokens.}
    \label{fig:aie_real_ab}
\end{figure*}

We previously mentioned using a context template for each relation for simplicity and complete control over the context. We conducted the first and second experiments using the actual retrieved document by the \textsc{Atlas} model's retriever component, as depicted in Figure~\ref{fig:aie_real_ab}. In the first experiment, object tokens continue to play a dominant role in the context as part of the translation process, where the model transforms object token representations from the encoder to the decoder for output generation (Figure~\ref{fig:aie_real_a_None}--\ref{fig:aie_real_a_attn}). In the second experiment, subject tokens demonstrate a significant impact on context relevance evaluation (Figure~\ref{fig:aie_real_b_None}--\ref{fig:aie_real_b_attn}), helping the model decide whether the context is suitable for extracting object tokens.

\end{document}